\pdfoutput=1

\documentclass[11pt]{article}

\usepackage[final]{naacl2021}

\usepackage{times}
\usepackage{latexsym}
\usepackage{hhline}
\usepackage[ruled,vlined]{algorithm2e}
\usepackage{graphicx}
\usepackage{multirow}
\usepackage{enumitem}
\usepackage{xcolor}
\usepackage{adjustbox}

\usepackage[T1]{fontenc}

\usepackage[utf8]{inputenc}

\usepackage{microtype}

\title{Graph-based Multilingual Product Retrieval\\ in E-Commerce Search}

\author{Hanqing Lu \\
        Amazon Search \\
        \texttt{luhanqin@amazon.com}
\And
        Youna Hu \\
        Amazon Search \\
        \texttt{ynhu@amazon.com}
\And
    Tong Zhao \\
    Amazon Personalization \\
    \texttt{zhaoton@amazon.com}
\AND
    Tony Wu\\
    Amazon Search \\
    \texttt{tonywu@amazon.com}
\And
    Yiwei Song \\
    Amazon Search \\
    \texttt{ywsong@amazon.com}
\And
 Bing Yin \\
 Amazon Search\\
 \texttt{alexbyin@amazon.com}
 }

\begin{document}
\maketitle
\begin{abstract}
Nowadays, with many e-commerce platforms conducting global business, e-commerce search systems are required to handle product retrieval under multilingual scenarios. Moreover, comparing with maintaining per-country specific e-commerce search systems, having a universal system across countries can further reduce the operational and computational costs, and facilitate business expansion to new countries. 
In this paper, we introduce a universal end-to-end multilingual retrieval system, and discuss our learnings and technical details when training and deploying the system to serve billion-scale product retrieval for e-commerce search. In particular, we propose a multilingual graph attention based retrieval network by leveraging recent advances in transformer-based multilingual language models and graph neural network architectures to capture the interactions between search queries and items in e-commerce search. Offline experiments on five countries data show that our algorithm outperforms the state-of-the-art baselines by 35\% recall and 25\% mAP on average. Moreover, the proposed model shows significant increase of conversion/revenue in online A/B experiments and has been deployed in production for multiple countries.

\end{abstract}

\section{Introduction}
Modern e-commerce search engines~\cite{DBLP:conf/kdd/HuangSSXZPPOY20, DBLP:conf/kdd/NigamSMLDSTGY19} typically consist of a retrieval stage and a ranking stage. The retrieval stage is responsible for collecting a set of relevant products with minimum computational resources. The ranking stage then applies sophisticated machine learning (ML) algorithms to determine their impression positions. 
Traditional retrieval models rely on keyword matching \cite{manning2008introduction}, which may lead to poor results when the exact term match is unavailable. Recently,  semantic matching models \cite{DBLP:conf/cikm/HuangHGDAH13,DBLP:journals/corr/PangLGXWC16} have been adopted to improve retrieval performance \cite{mitra2018introduction}. These models are trained using click/purchase logs and typically separated by countries \cite{DBLP:conf/wsdm/AhujaRKSR20}. However, such per-country specific training schema exposes three major drawbacks. First, maintaining country-specific models increase both operational burden and model iteration risks among countries. Second, the small amount of training data in low traffic countries may limit the ML model performance and this can also block the business expansion to new countries. Third, such models can not handle second language searches well. For example, the training data in US are dominated by English, which produces a model that cannot handle Spanish searches well. To solve above issues, ideally, a multilingual semantic retrieval model should be considered over monolingual retrieval models. However, how to design an effective and scalable multilingual semantic retrieval model for industry grade e-commerce search engine remains unsolved.

Built upon the success of pre-trained transformer-based models \cite{devlin2018bert, DBLP:journals/corr/abs-1906-08237, liu2019roberta} such as Bidirectional Encoder Representations from Transformers (BERT) 
for natural language processing, multilingual BERT (M-BERT) has also demonstrated success for multilingual tasks \cite{DBLP:conf/acl/PiresSG19}. Though the techniques are promising, it is not straightforward to directly apply them to our problem due to the \textit{vocabulary gap} issue \cite{mandal2019query}, i.e., customer searched queries are often short and from spoken input (e.g. `fancy clothes') but product descriptions are usually in formal written style (e.g. `formal attire'). There lacks a well established practice for fine-tuning multilingual BERT models on product search retrieval tasks. 

In this work, we address the vocabulary gap by sharing information between queries and products in the model via graph convolution networks (GCN). 
The query-to-product purchase/click logs naturally form a bipartite graph where each clicked product links with searched queries as neighbors. We expect to improve the product representation for retrieval tasks by incorporating information from its neighbor queries. 
For example, it is difficult for neural networks to directly match the query `great gifts for child' to the product `Disney puzzles for kids' given the vocabulary gap. But using the information that the product is connected with query `children gifts', we can incorporate this information in its final representation, and the product will have a higher chance to be matched with the given query. 

This paper presents an end-to-end multilingual retrieval system for e-commerce search engine. Our contributions are three-fold. \textbf{Model:} We present a general framework that is compatible with any transformer-based models and any GCN architectures to capture interactions between products and search queries; \textbf{Practice:} We provide a principled and practical guide of how to train the proposed model for large-scale product retrieval problem, e.g., how to define effective loss functions, how to feed online model-based hard negative samples to train the model and how to train the multilingual model with a novel one-language-at-a-batch (Sec \ref{sec:data_fusion}) approach; 
\textbf{System:} We discuss how to deploy the model to support product retrievals in multiple countries for e-commerce search.

To validate the effectiveness of our proposed method, we take offline experiments on billion-scale data across five languages and conduct online A/B testing experiments to measure the real traffic impacts. Through experimental results, our model outperforms state-of-the-art baselines by more than 25\% and increases revenue and conversion over the current production system.

\section{Methodology}
\label{sec:methodology}

We formulate the search retrieval task as follows. Supposed that we have a set of products $P = \{p_1, ..., p_n\}$. Each product $p_i$ has a number of neighbor queries $Q_i = \{q_{i,1}, ..., q_{i,t}\}$ where $(q_{i,j}, p_i)$ appears in the search logs (customers search for $q_{i,j}$ and purchase $p_i$). For an arbitrary query $q$, we want to find the top-$K$ relevant products from $P$. Note that $q$ is not in $Q = \{Q_1, ..., Q_n\}$ when our retrieval system handles unseen queries.
\begin{figure*}[htbp]
\centering
\includegraphics[width=0.95\textwidth]{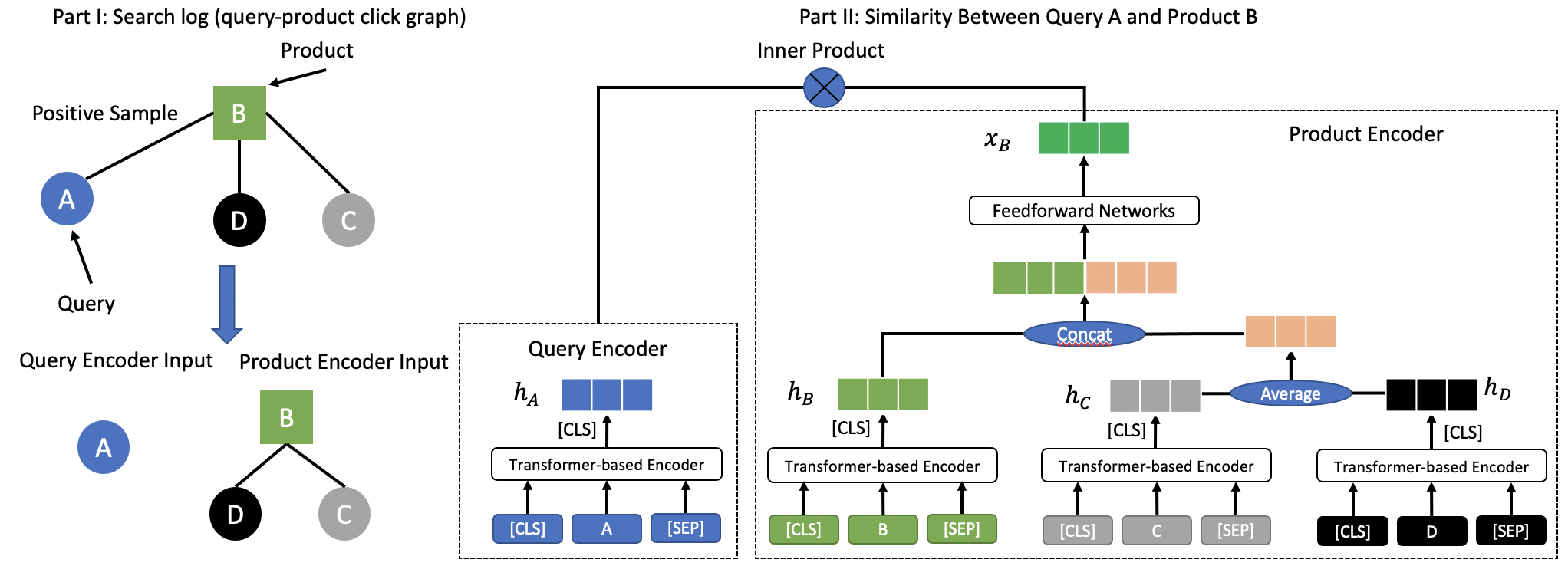}
\caption{The base architecture of multilingual GCN. Each circle indicates a query and each rectangle represents a product. Query $A$ and product $B$ is a positive pair in the training data, while $C$ and $D$ are the neighbor queries of $B$. The query encoder takes query $A$ as the input and outputs $A$'s embedding, $h_A$. The product encoder takes $B$, $C$, $D$ as the input and output $B$'s embedding, $x_B$.}
\label{fig:model}
\vspace{-6mm}
\end{figure*}
\subsection{Model Architecture}
The model has two main components: (1) a query encoder that encodes search queries; (2) a product encoder that encodes both the product description and its neighbor queries. The product encoder has a GCN component that encapsulates the neighbor queries and product information. 

\textbf{Query Encoder}:
The query encoder could be any transformer-based encoder, such as BERT \cite{devlin2018bert}, XLNet \cite{DBLP:journals/corr/abs-1906-08237}, DistilBERT \cite{sanh2019distilbert}, and RoBERTa \cite{liu2019roberta}.  Choosing these transformer-based encoders over other encoders (e.g. LSTM \cite{hochreiter1997long}) has several benefits. First, these models employ the word-piece tokenization which is robust to spelling errors and allows us to share vocabulary between different languages. In addition, transformer-based models can be easily parallelized and deployed online. We use the last hidden states of the encoders' [CLS] token as the embeddings for the query. For the other computationally more costly option of using the average pooling of the last hidden states of all tokens, we did not observe significant performance difference. 

\textbf{Product Encoder:}
The product encoder consists of 1) a transformer based encoder layer to extract the features of a product and its neighbor queries, and 2) a graph convolution layer that aggregates the extracted features to compute the final embeddings for a given product. The transformer-based encoder layer shares its parameters with the query encoder, and we also use the last hidden states of the encoder's [CLS] token as the features. The operations in the graph convolution layer are described in Algorithm \ref{gcn}.

\begin{algorithm}
\small
\textbf{Input:} extracted feature $\mathbf{h}_{p_i}$ of a product $p_i$, extracted features $\{\mathbf{h}_{q_{i,1}}, ..., \mathbf{h}_{q_{i,t}}\}$ of $p_{i}$'s neighbor queries $\{q_{i,1}, ..., q_{i,t}\}$\\
\textbf{Output:} the final product embedding $\mathbf{x}_{p_i}$ \\
\textbf{Step 1:} $\mathbf{h}_{q_i} = \frac{1}{t} \sum_j{\textrm{ReLU}(\mathbf{W_q} \cdot \mathbf{h}_{q_{i,j}}} + \mathbf{b_q}$) \\
\textbf{Step 2:} $\mathbf{x}_{p_i} = \textrm{ReLU}(\mathbf{W_p} \cdot \textrm{CONCAT}(\mathbf{h}_{p_i}, \mathbf{h}_{q_i}) + \mathbf{b_p})$
\caption{Graph Convolution Layer}
\label{gcn} 
\end{algorithm}

The intuition of adding the graph convolution layer to the product encoder is that it can fill the vocabulary gap between queries and product descriptions. With the vocabulary gap and length distribution discrepancy between queries and product descriptions, directly using the transformer-extracted embeddings of queries and products for matching is sub-optimal. By incorporating neighbor query information into the product representation, the matching model not only learns information from query-to-product similarity, but also learns from query-to-query similarity. This is especially helpful to tail queries that have limited behavior signals. 

Note that our framework is compatible with any GCN architectures in theory, such as GCN \cite{DBLP:conf/iclr/KipfW17} and GAT \cite{Petar2018graph}, and we will leave those explorations for future work.

\textbf{Loss Function:}
In the training stage, we use a pairwise ranking loss to train the model. Specifically, for each query $q_i$ in the training set, we sample a positive product $p_{i+}$ and a negative product $p_{i-}$. The triplet loss is defined as
\begin{equation}
\small
    L = \sum_i \mathrm{Log}(1 + \mathrm{exp} (\mathbf{x}_{q_{i}} \cdot \mathbf{x}_{p_{i-}} - \mathbf{x}_{q_i} \cdot \mathbf{x}_{p_{i+}}))
\end{equation}
The intuition is that 
we want the inner product of the positive pairs <$\mathbf{x}_{q_i}$, $\mathbf{x}_{p_{i+}}$> to be larger than the inner product of the negative pairs <$\mathbf{x}_{q_i}$, $\mathbf{x}_{p_{i-}}$>and the margin to be as large as possible.

\subsection{Training Details}
\label{training scheme}
There are two key factors in successfully training the aforementioned framework: 1) how to select proper negative samples for training the model; 2) how to properly feed training data from different languages to the model.

\textbf{Negative Sampling}: 
\label{sec:Negative_mining} 
Defining negative samples for semantic retrieval tasks is a tricky problem. A widely-adopted method is \textit{random sampling}, where one can randomly sample a product from the product catalog as the negative samples for a given query. However, this simple setting would generate sub-optimal results, since the randomly selected negative samples can be too easily distinguished from the positive samples.
Thus, it rarely brings knowledge for model learning and produces a model with low discriminative power for the retrieval task. 

Recent research work has indicated that using hard negatives could improve the model performance for retrieval tasks \cite{ying2018graph, DBLP:conf/kdd/NigamSMLDSTGY19, DBLP:conf/kdd/HuangSSXZPPOY20}. The hard negative sample should be the product that is somewhat related to the query but not a exact match. In the search retrieval scenario, we can define the following three kinds of hard negatives. \\
\textit{Behavior-based hard negative}: the negative samples are defined by users' behavior and are extracted from the search logs. For a given query, we take those products that were shown to the customer but not clicked as the hard negatives. \\
\textit{Offline model-based hard negative}: the negative samples are calculated by the current model in an offline fashion. Specifically, we first use the current model to generate the embeddings for all queries and products in the training set, and then calculate the inner product between all queries and all products. For each query, we sample negatives from its top-200 to top-1000 relevant products.\\
\textit{Online model-based hard negative}: the negative samples are generated on-the-fly with model learning. Specifically, we first randomly sample a batch of products. Then we use the current model to calculate the inner product between embeddings of these products and a batch of queries. For each query in the batch, we select the product with highest inner product value as the hard negative sample. 

We argue that the online model-based hard negative is the most suitable sampling method to our application. The behavior-based hard negative samples requires additional data collection processes and often yields worse results in the search retrieval task \cite{DBLP:conf/kdd/HuangSSXZPPOY20}. In fact, it is more suitable to the ranking task where the candidate pool is more refined. Besides, the offline model-based hard negative is too time-consuming, as we have to compute K-NN for each query in training set when we select/update the hard negatives. 

\textbf{Multilingual Data Fusion}: 
\label{sec:data_fusion}
How to properly feed the multilingual data to the model is another crucial factor to the training process. The amount of training data from different languages/countries varies greatly, and therefore low-resource languages would be underrepresented in the neural network model.
Inspired by \cite{devlin2018bert}, we perform exponentially smoothed weighting of the data. We would take the exponent of the percentage of  a language by factor $S$ and then re-normalize.
Suppose there are two languages, English and Spanish, which accounts for 90\% and 10\% of training data respectively. The re-normalized distribution is 
$\frac{{0.9}^{0.7}}{{0.9}^{0.7} + {0.1}^{0.7}} = 0.82$
for English. Therefore, high-resource languages will be under-sampled, and low-resource languages will be over-sampled. 

We also find that mixing training samples from multiple languages in one training batch makes it harder to train the model. Firstly, the negative sampling space is more complicated: we could sample a Spanish product as the negative sample of a English query. These easy negatives provide little knowledge to the model. In addition, different languages of training data appear in the same batch, which makes the batch gradient less stable. We propose to train the model with \textit{one-language-at-a-batch}, and make the negative sampling process language-dependent. In the experiment, we observe that doing so dramatically increases the performance on all languages by 5-6\% recall.

\section{Deployment}
The deployment of the proposed model has two parts: a query encoder and pre-computed product embeddings. As the query encoder is a standard transformer-based model and many papers have talked about the serving of it, this part can be easily deployed online. For the pre-computed product embeddings, we first compute the embeddings for products in our catalog, and incrementally update the product embedding periodically. To avoid repeated computations during the inference time (multiple products might have the same neighbor query), we first use the transformer-based encoder to compute the embeddings for all queries and products in the graph, and then join the products' embeddings with their neighbor queries' embeddings. Lastly, we pass these intermediate embeddings through the GCN layer to generate the final product embeddings. During the search retrieval step, we simply use the query encoder to extract the embedding for an input query. Then, we find the K-nearest neighbors (K-NN) products by calculating the cosine-distance between the query embedding and the pre-computed product embeddings. The K-NN products are used to augment the matchset of the given query. 

\begin{table*}[tbp]
\footnotesize
\small
\caption{Matching performance for our model and baselines.}
\centering
 \vspace{-3mm}
\label{tab:baseline}
\begin{adjustbox}{center}
\begin{tabular}{|c|c|c|c|c|c|c|c|c|c|c|}
\hline
    & \multicolumn{2}{c|}{US} & \multicolumn{2}{c|}{ES} & \multicolumn{2}{c|}{FR} & \multicolumn{2}{c|}{IT} & \multicolumn{2}{c|}{DE}\\ 
\hline
  method & recall          & mAP             & recall          & mAP  &  recall          & mAP  &  recall      & mAP &  recall          & mAP    \\
\hline
DSSM & 49.53\%  & 34.09\%    & 38.82\%  & 22.26\% &  38.16\% &  21.98\%   &  42.51\% & 24.68\% & 46.87\%  &  30.16\% \\
Multilingual BERT &  38.82\%  &  25.14\%   & 23.06\%  & 12.07\%  &  25.41\%  & 13.67\% & 24.36\% & 13.06\% & 25.31\%  &  15.18\%  \\
Our model w/o BERT   & 79.79\%  & 60.06\%    & 66.53\%  & 39.01\% & 68.01\%  & 41.43\% & 70.32\%  & 42.66\% & 74.69\%  &   51.79\%     \\
Our model w/o GCN    & 80.83\%  &  60.68\%  & 68.98\%  & 41.73\% & 70.03\%  & 42.28\%   & 72.88\% & 44.98\% & 76.69\%  & 53.75\%  \\
Our model   &  \textbf{85.86\%} &  \textbf{66.69\%} & \textbf{73.60\%}  & \textbf{44.40\%} & \textbf{74.97\%}  &  \textbf{47.07\%}    & \textbf{77.16\%}   & \textbf{48.03\%} & \textbf{81.44\%}  & \textbf{58.33\%} \\
\hline
\end{tabular}
\end{adjustbox}
\vspace{-3mm}
\end{table*}

\section{Experiments}
\label{experiments}
We collect the data from a large e-commerce platform that has business in multiple countries. To provide a comprehensive understanding on the role of multilingual queries in a real-world product search system, we select five countries: United States (US), Spain (ES), France (FR), Italy (IT) and Germany (DE). We subsample our data from one year of search log in each country. We organize the collected search log into query-product pairs with different customer behavior signals, e.g. click/purchase.
For model offline testing, we first randomly sample 20K queries from each country. We then use our algorithm to rank a sub-corpus of 100K products (in each country) for those queries. The 100K product corpus consists of purchased products for those 20K queries and additional random negatives. Since our work focuses on the retrieval part of a product search engine, we adopt two matching metrics to summarize our results: Recall@10 (recall) and mean Average Precision (mAP). We employ the multilingual DistilBERT \cite{sanh2019distilbert} with 6 layers and 768 hidden units as our encoder. We set the batch size to 640 and use Adam optimizer \cite{kingma2014adam} with $\alpha = 0.0001, \beta_1=0.9$, and $\beta_2=0.999$. We run all the experiments on an AWS p3dn.24xlarge instance with 768GB memory and 8 NVIDIA V100 GPUs. We train the model on 8 GPUs in a distributed fashion. The model is trained for 140K batches, where the 28K `warm up' batches are trained with random negatives and remaining batches are trained with online model-based hard negatives.

\subsection{Comparison Results}
We compare against the following baselines:\\
\textbf{DSSM} \cite{DBLP:conf/cikm/HuangHGDAH13} is an earlier work to extract the semantic representations of queries and documents from large-scale click-through data by leveraging deep neural networks. We train 5 language-specific DSSM models with monolingual training data. \\
\textbf{Multilingual BERT} \cite{devlin2018bert} is the vanilla BERT without any fine-tuning. We use the \textit{bert-base-multilingual-cased} 
model from huggingface implementation. We directly use the Multilingual BERT to encode the queries/products, and take the output [CLS] embeddings as the representations of queries/products. \\
\textbf{Our model w/o BERT} is a variant of our model, where we replace the DistilBERT encoder with a one-layer feed forward neural network and use the same word embedding matrix as the DistilBert. We train this model with exactly the same settings as we train the main model. \\
\textbf{Our model w/o GCN} is a variant of our model, where we remove the GCN module. It means that we only use product descriptions to get the product embeddings, and there is no graph convolution layer in the product encoder.
Table \ref{tab:baseline} shows (1) Multilingual BERT without any fine-tuning does not work for multilingual search retrieval tasks. It has the lowest recall and mAP, which proves the necessity of designing proper fine-tuning tasks for BERT-based model; (2) replacing the feed forward neural networks with DistilBERT leads to 6\% - 7\% recall and 5.5\% - 6\% mAP improvement on all languages; (3) adding GCN module to the product encoder further achieves significant boosts (5-6\% recall improvement and 4.5\%-6\% mAP improvement), suggesting that GCN help the \textit{vocabulary gap} issue.

\subsection{Ablation Study}
\textbf{Negative Sampling:} 
We try three kinds of hard negatives as illustrated in Section \ref{sec:Negative_mining}.
Table \ref{tab:hardnegative} shows the results of training our model with different hard negatives. We observe that the performance of offline model-based hard negatives is similar to that of online model-based hard negatives (< 0.4\% recall difference). However, computing the offline model-based hard negatives takes a total of 4x training time. Besides, training with behavior-based hard negatives has worst results (-10\% recall, -7\% mAP), because behavior-based negatives are not appropriate for retrieval tasks \cite{DBLP:conf/kdd/HuangSSXZPPOY20}, since most impressed products are often relevant to the query. Including them as negatives confuses the model from focusing on retrieval tasks.

\textbf{Multilingual Training Data Fusion:}
We test three data fusion strategies: 1) sample by unweighted data size + train with one language at a batch (\textit{unweight+separate}); 2) sample by exponentially weighted data size + train with mixed languages in a batch (\textit{weight+mix}); 3) sample by exponentially weighted data size + train with one language at a batch (\textit{weight+separate}). 
Table \ref{tab:multilingual} shows the results from different multilingual data fusion strategies. By exponentially weighting the training data, we can improve the matching performance in low-resource languages (ES, FR, IT) without hurting the performance in high resource languages (DE and US). Besides, \textit{weighted+separated} beats \textit{weighted+mixed} by 2-3\% recall and 1-2\% mAP margin on all languages, suggesting that training with one-language-at-a-batch is superior to mixed training.

\begin{table*}[tbp]
\footnotesize
\caption{Matching performance with different kinds of hard negatives.}
 \vspace{-3mm}
\label{tab:hardnegative}
\begin{adjustbox}{center}
\begin{tabular}{|c|c|c|c|c|c|c|c|c|c|c|}
\hline
    & \multicolumn{2}{c|}{US} & \multicolumn{2}{c|}{ES} & \multicolumn{2}{c|}{FR} & \multicolumn{2}{c|}{IT} & \multicolumn{2}{c|}{DE}\\ 
\hline
  hard negative & recall          & mAP             & recall          & mAP  &  recall          & mAP  &  recall      & mAP &  recall         & mAP    \\
\hline
behavior & 76.62\% & 59.36\% & 63.28\%  & 38.80\% & 63.90\%  & 40.31\%    & 66.17\%  & 42.18\% & 68.44\%  &  49.67\%      \\
 offline model &  85.44\% &  66.43\%    & \textbf{73.95\%}   &  \textbf{44.33\%} & 74.86\%  & 46.76\%    &\textbf{77.43\%}    &\textbf{48.20\%} &  \textbf{81.52\%}  &  \textbf{58.68\%}   \\
online model & \textbf{85.86\%} &  \textbf{66.69\%} & 73.60\%  & 44.40\% & \textbf{74.97\%}  &  \textbf{47.07\%}    & 77.16\%   & 48.03\% & 81.44\%  & 58.33\%       \\
\hline
\end{tabular}
\end{adjustbox}
\end{table*}

\begin{table*}[tbp]
\footnotesize
\caption{Matching performance with different multilingual data fusion strategies.}
 \vspace{-3mm}
\label{tab:multilingual}
\begin{adjustbox}{center}
\begin{tabular}{|c|c|c|c|c|c|c|c|c|c|c|}
\hline
& \multicolumn{2}{c|}{US} & \multicolumn{2}{c|}{ES} & \multicolumn{2}{c|}{FR} & \multicolumn{2}{c|}{IT} & \multicolumn{2}{c|}{DE}\\ 
\hline
fusion strategy  &recall          & mAP             & recall          & mAP  &  recall          & mAP  &  recall      & mAP &  recall          & mAP    \\
\hline
 weight+mix &  84.61\% & 65.24\% & 70.45\%  & 42.01\% & 71.41\%  & 44.34\%    & 73.39\%  & 45.46\% & 78.85\%  &  55.91\%      \\
 unweight+separate &  85.82\% &  66.60\%    & 71.69\%   &  42.93\% & 73.29\%   & 45.63\%    &74.53\%    &46.66\% &  80.61\%  &  57.73\%   \\
weight+separate & \textbf{85.86\%} &  \textbf{66.69\%} & \textbf{73.60\%}  & \textbf{44.40\%} & \textbf{74.97\%}  &  \textbf{47.07\%}    & \textbf{77.16\%}   & \textbf{48.03\%} & \textbf{81.44\%}  & \textbf{58.33\%} \\
\hline
\end{tabular}
\end{adjustbox}
\end{table*}

\subsection{Online Experiments}
We report our findings from online A/B experiments on a large-scale e-commerce website with our multilingual GCN model. We run online match set augmentation experiments in three countries and two languages. The proposed algorithm significantly improves business metrics in all countries, leading to +1.8\% increase in average clicks, +0.3\% in revenue, and +0.4\% in conversion. We also observe reformulated searches decreased by 1\%. This reduction results in customers finding their desired products with less effort, likely from that our model bridges the vocabulary gap between queries and products. All results provide evidence that our algorithm leads to better retrieval performance and can help customers fulfill their shopping missions.

\section{Related Works}
Search engine retrieval has been based on lexical match to identify relevant documents for queries. Recently, major industry search engines \cite{DBLP:conf/kdd/NigamSMLDSTGY19, DBLP:conf/kdd/HuangSSXZPPOY20, DBLP:conf/kdd/FanGZMSL19} have incorporated \textit{semantic matching} for improvements. Such algorithms can be classified into \textit{embedding-based models} and \textit{interaction models}. Embedding based models such as DSSM \cite{DBLP:conf/cikm/HuangHGDAH13} and its subsequent works \cite{shen2014latent,palangi2016deep,hu2014convolutional} convert queries and documents into embeddings for retrieval. Interaction models, like MatchPyramid \cite{DBLP:journals/corr/PangLGXWC16} and DRMM \cite{guo2016deep} leverage interaction matrices to capture local term matching. However, they are computationally costly for industry data. 


With BERT \cite{devlin2018bert} becoming the state-of-the-art embedding method, it is adopted for various applications \cite{yang-etal-2019-enhancing-pre, yu-etal-2020-named, DBLP:conf/sigir/KhattabZ20, DBLP:conf/iclr/HumeauSLW20, DBLP:conf/iclr/ChangYCYK20}. However, how to properly fine-tune BERT for retrieval tasks in product search remains unstudied. Our work fills this gap and provides a practical guide to fine-tune BERT-based models using production-scale search data. Furthermore, M-BERT provides representations for 104 languages and has proven ability to handle multilingual tasks \cite{DBLP:conf/acl/PiresSG19}. Other multilingual embedding models have also been proposed and validated \cite{DBLP:conf/rep4nlp/SchwenkD17, DBLP:conf/nips/ConneauL19, DBLP:conf/acl/ConneauKGCWGGOZ20}. Our method is flexible and so that all these models can serve as a component. 


Notably, our multilingual problem is different from the cross-lingual information retrieval (CLIR) problem~\cite{DBLP:series/synthesis/2010Nie, DBLP:conf/lrec/JiangEHKZ20} ,  
which refers to the scenario where the query is in one language but document is in other languages. 
In our problem, product descriptions and search queries are always in the same primary language and except a small fraction in different languages


Graph neural network is gaining prominence in ML applications \cite{ying2018graph, zhang2019neural}. The notion of "graph convolutions" is first proposed in \cite{bruna2013spectral} with spectral graph theory. Later, GraphSAGE \cite{NIPS2017_6703} redefines it to avoid operating on the entire graph. Recent efforts \cite{wang2019neural, berg2017graph} adopt GCN to the user-item interaction graph and leverage the neighbors for recommendation. 
LightGCN \cite{DBLP:conf/sigir/0001DWLZ020} reported that neighborhood aggregation is the only important component of GCN, and weighted-sum of neighbor embeddings yield the best results. Our method leverages GCN to incorporate neighbor queries' information into product embedding, which bridges the vocabulary gap between query and product. Moreover, our framework is compatible with any GCN architectures, so can leverage the advances there.

\section{Conclusion}
Our paper present a multilingual graph convolution networks model for language-agonistic semantic retrieval in product search engine. Our method not only can handle multilingual text data, but also addresses the \textit{vocabulary gap} issues between queries and product descriptions. We also provide a practical guide of fine-tuning the proposed model on retrieval tasks. 
We conduct various experiments including offline evaluation on 5 languages and online A/B test in three countries. In all experiments, our model consistently beats the baselines and demonstrates improved product discoverability.

\bibliography{naacl2021}
\bibliographystyle{acl_natbib}
\end{document}